\documentclass[acmtog]{acmart}
\AtBeginDocument{%
  }

\usepackage{multirow}

\begin{document}

\title{High-Fidelity Single-Image Head Modeling with Industry-Grade Topology}

\author{Yunmu Wang}
\authornote{The first three authors contributed equally.}
\author{Zoubin Bi}
\authornotemark[1]
\author{Bowen Cai}
\authornotemark[1]
\author{Chenchu Rong}
\author{Jinlong Wang}
\author{Junchen Deng}
\author{Aocheng Huang}
\author{Jidong Jia}
\author{Huan Fu}
\affiliation{%
  \institution{Alibaba Group}
  \city{Beijing}
  \country{China}
}

\renewcommand{\shortauthors}{Wang et al.}

\newcommand{\figref}[1]{Fig.~\ref{#1}}
\newcommand{\tabref}[1]{Tab.~\ref{#1}}
\newcommand{\appref}[1]{Appendix~\ref{#1}}
\newcommand{\stepref}[1]{Step~\ref{#1}}
\newcommand{\eqnref}[1]{Eq.~\ref{#1}}
\newcommand{\algref}[1]{Algorithm~\ref{#1}}
\newcommand{\note}[1]{{\color{blue}#1}}
\newcommand{\warning}[1]{{\color{red}#1}}
\newcommand{\link}[1]{{\color{magenta}#1}}
\newcommand{\secref}[1]{Sec.~\ref{#1}}

\begin{abstract}
We present a \textbf{single-image} head mesh reconstruction framework that addresses the longstanding challenge of simultaneously preserving \textbf{facial identity} and producing \textbf{industry-grade} topology.
Our framework adopts a coarse-to-fine optimization pipeline that refines a rigged template across three stages---rig, joint, and vertex---achieving stable convergence and consistent topology.
To mitigate the ill-posed nature of single-image 3D face reconstruction and ensure identity preservation, we employ a normal consistency objective jointly with landmark alignment.
To further preserve local surface structure and enforce topological regularity, we introduce geometry-aware constraints based on Gaussian curvature and conformal consistency, along with auxiliary regularizations that correct fine artifacts such as lip seams and eyelid discontinuities.
Our hierarchical optimization with geometry-aware regularization yields meshes with semantically meaningful edge flow and industry-grade topology.
After geometry reconstruction, we extract UV-space texture and normal maps to preserve appearance details for visualization and downstream use.
In a user study with \textit{22} professional technical artists, our results were assessed as approaching industry-grade usability, and \textbf{95\%} of participants ranked our method as the top-performing approach, underscoring its effectiveness for real-world digital human production.
\end{abstract}

\begin{teaserfigure}
    \centering
    \begin{minipage}{\textwidth}
        \includegraphics[width=\linewidth]{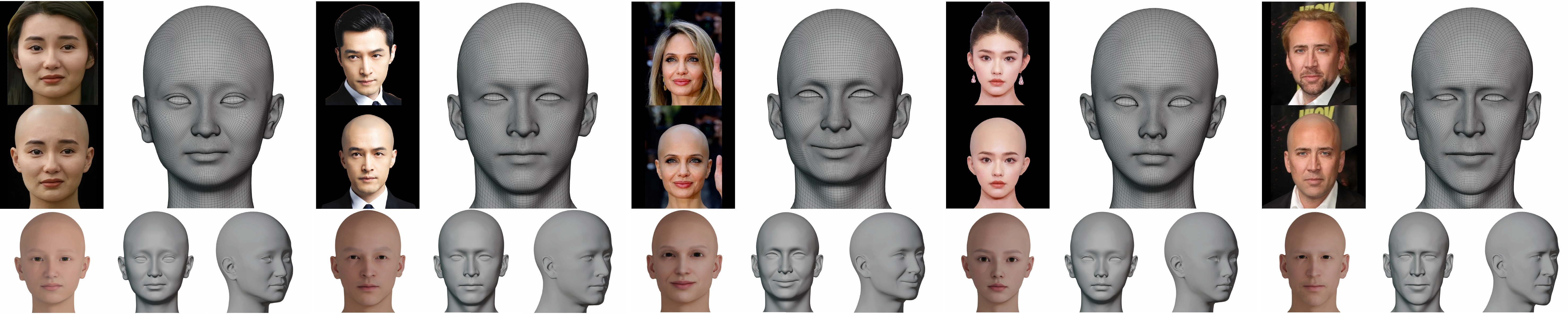}
    \end{minipage}
    \vspace{-0.3cm}
    \caption{\textbf{Single-image head modeling results of our method.} Our method generates high-fidelity, industry-grade head meshes with clean topology that can enter downstream binding and animation workflows. Frontalized and de-occluded inputs are shown for reference. Additional results are provided in the accompanying video and supplemental material. Please zoom in to inspect fine details.
    }
    \label{fig:teaser}
\end{teaserfigure}



\maketitle
\section{Introduction}
\label{sec:intro}

High-fidelity head meshes play a central role in many vision and graphics applications, ranging from digital human production to AR/VR and digital twins. In professional film and VFX pipelines, a usable head mesh must satisfy more than visual realism. It should have clean topology, anatomically faithful geometry, and semantically consistent edge flow in critical deformation regions such as the eyelids and lips. It must also be ready for downstream binding with minimal manual cleanup. Producing such assets still depends largely on skilled artists using specialized tools, making the process labor-intensive and often requiring hours or days for each individual head. This manual bottleneck limits scalability and consistency, motivating research on automated, high-quality head modeling.

The central challenge in automating digital head creation lies in generating a topologically consistent mesh that adheres to artist-defined edge layouts while faithfully preserving identity and fine-scale geometry (e.g., wrinkles and eye contours). Existing approaches fall into two main paradigms: 3D Morphable Models (3DMMs) and 3D registration. 3DMMs~\cite{blanz2023morphable} offer a compact and controllable representation of facial geometry, enabling stable reconstruction from limited inputs. Although larger datasets and richer basis spaces~\cite{paysan20093d,li2017learning} have improved their capacity, 3DMMs still struggle to reproduce identity-specific nuances and high-frequency realism. Conversely, the scan-to-registration paradigm, which forms the foundation for constructing 3DMM bases, remains the most reliable route for accurately digitizing real individuals. Despite advances in registration algorithms~\cite{zheng2022imface,vesdapunt2020jnr,yang2023asm}, substantial manual refinement by digital artists is still required to correct local artifacts, particularly around delicate regions such as the eyelids and lips.
For subjects that cannot be directly scanned, achieving a high-fidelity, cleanly organized head mesh remains an open challenge—even with the advent of large-scale generative models~\cite{hong2023lrm,zhang2024clay,wang2025vggt}.

In this work, we present an automatic optimization framework that reconstructs high-fidelity, industry-grade head meshes from a single facial image and can be instantiated on different rigged templates.
Our method deforms a standardized, rigged facial template in a coarse-to-fine manner—optimizing successively at the global rig, joint, and vertex levels—to progressively introduce semantically meaningful deformations, which enhance geometric accuracy while preserving clean topology.
The deformation process is driven by two key cues derived from the input image: a predicted normal map and 2D facial landmarks. Normals provide strong local geometric guidance and are less ambiguous than depth estimates in monocular settings. We tailor landmark supervision to different feature types: Mean Squared Error (MSE) loss for sparse corner points (eyes, mouth) and Chamfer Distance (CD) for dense contours (eye shapes), which better preserves their respective geometric properties. However, these cues alone can distort mesh quality and fine-scale topology. We therefore regularize the optimization with Gaussian curvature, which preserves intrinsic surface smoothness, and a conformal constraint that prevents local angular distortions. Additional lightweight priors are applied to correct subtle yet perceptually important defects such as lip seams and eyelid discontinuities.
Following geometry reconstruction, we capture texture and normal maps in UV space, while reusing the remaining material maps from the template to assemble the final digital head. The extraction procedure is straightforward, involving intrinsic decomposition, gradient-based normal computation, and Dual-Mask Poisson Blending.

Extensive experiments demonstrate that our method generates identity-preserving head meshes with clean, artist-defined topology. It achieves an ArcFace cosine similarity of \textit{0.3770} and a geometric RMSE of \textit{0.0159}, outperforming state-of-the-art single-image reconstruction approaches. Despite relying on only a single input image, our method attains geometric accuracy comparable to recent scan-based registration methods in our controlled evaluation. In a user study with \textit{22} professional technical artists, participants rated our topology as nearing industry-grade usability, and \textit{95\%} ranked our method as the top-performing approach. Together, these findings support the use of our reconstructions in downstream binding workflows and subsequent animation tests.

\section{Related Work}
\label{sec:rel}

\subsection{Reconstruction}
\noindent \textbf{Parametric models.}
Reconstructing 3D faces from limited observations is inherently ill-posed, as multiple shapes may correspond to the same 2D projection. Parametric methods address this ambiguity by constraining the solution within a low-dimensional latent space learned from 3D scans, imposing strong statistical priors on geometry and appearance.
The 3D Morphable Model (3DMM)~\cite{blanz2023morphable} represents shape and texture in a PCA space and has since been extended with larger datasets~\cite{paysan20093d,li2017learning,booth2018large} or in-the-wild images~\cite{kemelmacher2013internet,booth20173d,feng2021learning}, and further disentangled identity and expression through bilinear formulations~\cite{vlasic2006face,cao2013facewarehouse,bolkart2015groupwise,yang2020facescape}.
Recent 3DMM variants~\cite{li2017learning,ploumpis2019combining,xu2020ghum,yang2020facescape,chai2022realy} provide richer control over pose and expression but remain limited by their linear subspaces, which fail to capture fine geometric variations.
To increase flexibility, non-linear and hierarchical latent models~\cite{ranjan2018generating,tran2018nonlinear,bouritsas2019neural,dib2024mosar,zheng2022imface,giebenhain2023learning} leverage deep architectures to learn complex shape manifolds. However, their capacity is still bounded by available high-quality scan data, which are costly to acquire.
Several works~\cite{feng2021learning,wang2022faceverse,lei2023hierarchical} enhance local fidelity by predicting displacement or normal maps, though these may distort mesh regularity and degrade deformation consistency. Rendering-based approaches such as MICA~\cite{zielonka2022towards} use inverse-rendering supervision but remain sensitive to lighting ambiguities. SMIRK~\cite{retsinas2024smirk} alleviates this by replacing differentiable rendering with neural synthesis for more faithful expressions.
Recent formulations, e.g., FlowFace~\cite{taubner20243d} and Pixel3DMM~\cite{giebenhain2025pixel3dmm}, directly predict UV-space flow fields or dense normal maps for higher-frequency refinement.
Overall, parametric models ensure stable, semantically interpretable reconstruction but sacrifice local expressiveness. Our method mitigates this trade-off through hierarchical optimization that maintains topological regularity while recovering fine-scale details.

\noindent \textbf{Non-parametric models.}
In contrast, non-parametric methods directly infer 3D geometry—via vertex positions, depth, or implicit fields—allowing more flexible, high-fidelity reconstruction. Classical multi-view stereo (MVS) systems~\cite{ma2007rapid,ghosh2011multiview,beeler2011high} achieve accurate geometry but are computationally heavy and rely on controlled setups.
Recent volumetric methods~\cite{li2021topologically,bolkart2023instant} predict faces in feature volumes for parameter-free, near-real-time inference, though their tessellation irregularity limits animation.
Single-image regressors~\cite{sela2017unrestricted,richardson2017learning,zeng2019df2net} remove multi-view dependency but often overfit, yielding artifacts or inconsistent surfaces.
The advent of neural implicit representations~\cite{mildenhall2021nerf,wang2021neus,chan2022efficient,zhang20233dshape2vecset,kerbl20233d,bi2024gs3} and large-scale generative architectures~\cite{vaswani2017attention,ho2020denoising,lipman2022flow,peebles2023scalable,wang2025vggt} has enabled impressive 3D reconstruction and synthesis~\cite{hong2023lrm,qiu2025lhm,chen2025hart,chu2024generalizable,wu2025fastavatar,wang2023rodin,zhang2024clay,feng2025arm,xiang2025structured,lai2025hunyuan3d,li2025sparc3d}.
Despite their realism, these models usually output volumetric or point-based representations requiring post-processing to extract meshes, leading to inconsistent topology across subjects. Such inconsistency impedes rigging, animation, and semantic editing.
While non-parametric methods excel in visual fidelity, they remain unsuitable for reconstructing industry-grade head meshes with stable topology for downstream binding and animation workflows.
Our framework combines the stability of parametric modeling with the geometric precision of non-parametric approaches, producing consistent, high-fidelity head meshes from monocular input.

\begin{figure*}[!th]
    \centering
    \includegraphics[width=0.95\linewidth]{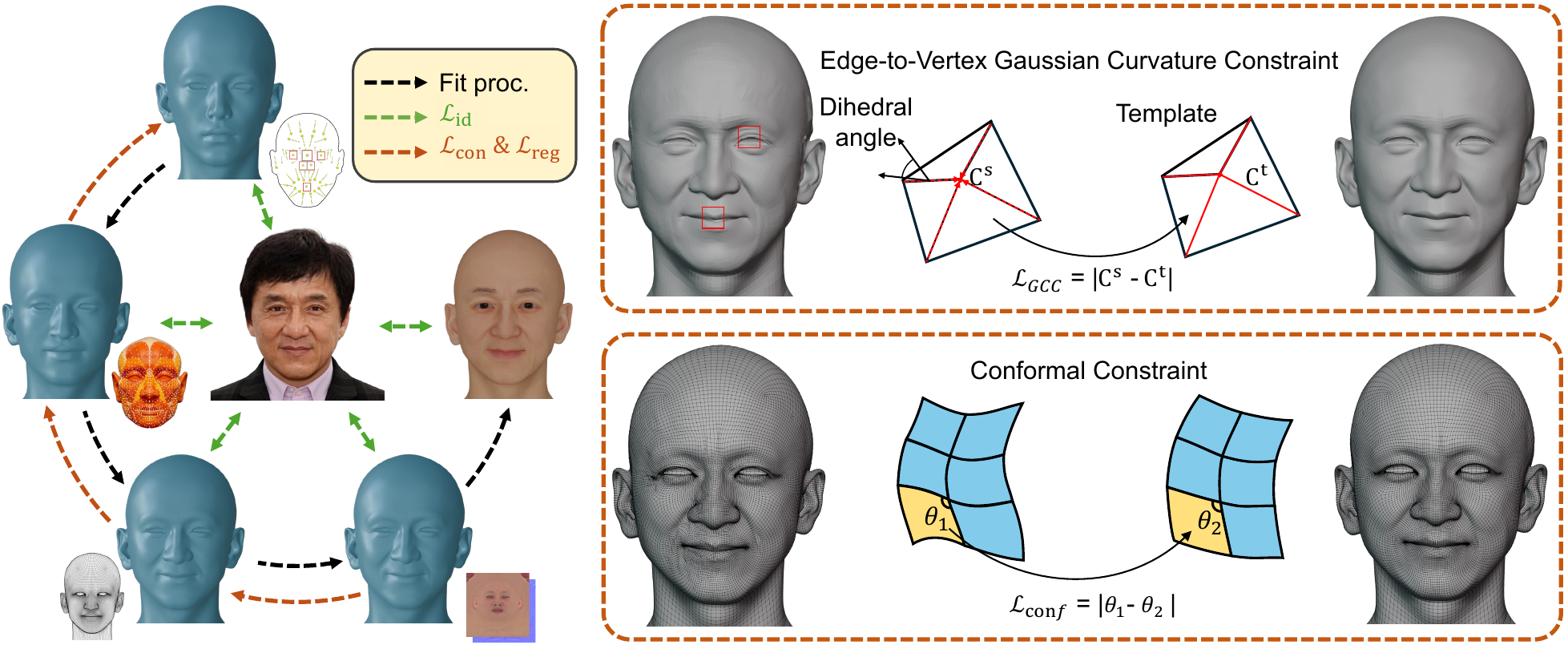}
    \vspace{-10pt}
    \caption{\textbf{Overview of our pipeline.}
    \textbf{Left:} Our method adopts a coarse-to-fine reconstruction framework that hierarchically deforms the template mesh from semantic structure to fine-scale details, yielding semantically meaningful edge flow and industry-grade topology. \textbf{Right:} We introduce Gaussian curvature and conformal consistency losses to simultaneously preserve local surface geometry and enforce topological regularity.
    }
    \label{fig:pipeline}
\end{figure*}

\subsection{Registration}
Registration aims to deform a template mesh to align with a scan or another irregular yet accurate mesh, producing high-detail surfaces with regular topology for animation and statistical modeling.
Classical non-rigid ICP methods optimize vertex transformations under local rigidity or smoothness priors~\cite{allen2003space,amberg2007optimal,sorkine2007rigid}, while extensions incorporate conformal~\cite{yoshiyasu2014conformal} or curvature-based~\cite{tajdari2022feature} regularization to preserve local shape characteristics.
Our approach similarly enforces conformality and curvature consistency but unifies them within a single optimization framework for locally faithful deformation.
Thin-plate spline (TPS) formulations~\cite{bookstein2002principal,chui2000new} and Coherent Point Drift (CPD)~\cite{myronenko2010point} provide alternative probabilistic or smooth functional mappings, though they remain computationally expensive or prone to underfitting in complex cases.
Gaussian Process Morphable Models (GPMMs)~\cite{luthi2017gaussian} generalize deformation priors within a probabilistic framework, later extended by adaptive skinning models (ASM)~\cite{yang2023asm} for improved accuracy with compact parameters.
Hierarchical strategies~\cite{cheng2017statistical,dai2018non,pears2023laplacian} improve convergence but typically rely on naive local scaling.
In contrast, our hierarchical parameterization leverages semantic structure for intuitive multi-level control.
More recent neural approaches, such as JNR~\cite{vesdapunt2020jnr} and SPHEAR~\cite{bazavan2024sphear}, learn compact neural representations or spherical embeddings to achieve efficient feedforward registration.
While registration-based pipelines rely on precise scan data and often require manual artifact correction, our method reduces this acquisition burden by reconstructing industry-grade head meshes from a single image, streamlining mesh creation for production workflows.

\section{Preliminary}
\label{sec:prelim}

\noindent \textbf{Rigged Template.}
We define a rigged template for a target mesh topology as
$\mathcal{T}=(\bar{\mathbf{V}}, \mathcal{F}, \mathcal{R}, \mathcal{J},
\bar{\boldsymbol{\theta}}, \mathbf{W}, \Delta)$,
where $\bar{\mathbf{V}}\in\mathbb{R}^{N\times 3}$ denotes the neutral vertex positions,
$\mathcal{F}$ is the fixed mesh connectivity, $\mathcal{R}$ is the set of rig controllers,
$\mathcal{J}$ is the joint hierarchy, $\bar{\boldsymbol{\theta}}$ denotes the default local
joint parameters, $\mathbf{W}$ contains the skinning weights, and $\Delta$ denotes residual vertex offsets. A different target topology requires a
corresponding rigged template defined under the same representation.

Following JNR~\cite{vesdapunt2020jnr}, ASM~\cite{yang2023asm}, and related rigged face
models, each joint $j_k\in\mathcal{J}$ is associated with a local transform relative to its
parent in the hierarchy. In addition to the joint layer, our template contains a controller
layer $\mathcal{R}$, similar in spirit to production facial rigs such as
MetaHuman~\cite{EpicGames2021Metahuman}. Each controller $r_i\in\mathcal{R}$ is a bounded
parameter associated with a semantically meaningful facial deformation. In this work, this
rig serves as a reconstruction-time deformation parameterization, rather than the downstream
facial rig used for expression animation.

Given a controller vector $\mathbf{r}$, the local joint parameters are driven by a sparse
controller-to-joint mapping,
$\boldsymbol{\theta}(\mathbf{r})=\bar{\boldsymbol{\theta}}+\Phi(\mathbf{r})$,
where $\Phi(\mathbf{r})$ aggregates controller-dependent offsets in joint-parameter space.
The corresponding joint transforms are obtained by forward kinematics over the hierarchy.
The deformed mesh is then computed by linear blend skinning with residual offsets:
\[
\mathbf{V}(\mathbf{r}) =
\mathrm{LBS}\bigl(\bar{\mathbf{V}}, \mathbf{T}(\boldsymbol{\theta}(\mathbf{r})), \mathbf{W}\bigr)
+ \Delta(\mathbf{r}),
\]
where $\mathbf{T}(\boldsymbol{\theta}(\mathbf{r}))$ denotes the hierarchy-consistent joint
transforms and $\Delta(\mathbf{r})\in\mathbb{R}^{N\times 3}$ denotes controller-dependent
residual vertex offsets. This formulation defines a three-level deformation hierarchy,
consisting of controller values, joint parameters, and vertex positions, which are optimized
successively in the following section.

\section{Methodology}
\label{sec:method}

\subsection{Framework}
Given the rigged template defined in \secref{sec:prelim}, our method reconstructs an \emph{industry-grade} head mesh from a \emph{single} image through hierarchical optimization. As illustrated in~\figref{fig:pipeline}, we successively optimize rig parameters, joint parameters, and vertex positions, corresponding to global deformation, mid-level articulation, and fine-scale geometric refinement. This coarse-to-fine strategy preserves clean topology and coherent edge flow:

\begin{enumerate}
\item \textbf{Rig-level optimization:} We first optimize the rig parameters, where each controller corresponds to a semantically meaningful facial region (e.g., the jaw, cheeks, or brows). The resulting deformations are propagated along the rig–joint–vertex hierarchy, enabling expressive and structurally consistent global shape changes.

\item \textbf{Joint-level optimization:} Next, we fix the rig and optimize the joint parameters, which retain local semantic meaning while offering finer local control. This stage refines mid-level geometry through localized deformations, ensuring a smooth transition from rig-driven deformation to dense surface reconstruction.

\item \textbf{Vertex-level optimization:} Finally, we directly optimize vertex positions to recover fine-scale surface details such as wrinkles and subtle skin folds, achieving high-fidelity reconstruction.
\end{enumerate}

By progressively refining from semantic to fine-scale deformations, our hierarchical optimization produces meshes that preserve individual identity comparable to registration-based methods while maintaining more coherent edge flow, achieving this high-fidelity reconstruction from only a single input image.

\subsection{Shape Reconstruction}

We illustrate our formulation using rig-level optimization as an example; the same objective generalizes across all three parameter levels, differing only in the optimized variables and hyperparameters. Given a \emph{single} input facial image $I$ and a rigged template $\mathcal{T}$ defined in \secref{sec:prelim}, we optimize the current parameter level to reconstruct the input identity while preserving topological consistency. The resulting mesh is evaluated through differentiable rendering with the unified objective
$
\mathcal{L} = \mathcal{L}_{\text{id}} + \mathcal{L}_{\text{con}} + \mathcal{L}_{\text{reg}},
$
where $\mathcal{L}_{\text{id}}$, $\mathcal{L}_{\text{con}}$, and $\mathcal{L}_{\text{reg}}$ denote the identity, topological consistency, and regularization losses, respectively.

In the following, we first introduce the primary supervision signals—normal and facial landmark constraints—that drive identity reconstruction (\secref{subsec:supervise}). We then present Edge-to-Vertex Gaussian Curvature and conformal constraints that enforce topological consistency (\secref{subsec:stiffness}). Finally, we describe additional regularization terms that address subtle yet critical artifacts (\secref{subsec:fine_artifacts}). 

\subsubsection{Facial Identity Restoration}\label{subsec:supervise}
To preserve high-frequency geometric details while mitigating the inherent ill-posedness of monocular 3D face reconstruction, our method integrates two complementary supervision signals:
$\mathcal{L}_{\text{id}} = \lambda_{\text{n}} \mathcal{L}_{\text{n}} + \lambda_{\text{land}} \mathcal{L}_{\text{land}},$
where $\mathcal{L}_{\text{n}}$ and $\mathcal{L}_{\text{land}}$ denote the normal and landmark supervision terms, and $\lambda_{\text{n}}$, $\lambda_{\text{land}}$ are their respective weights. Normal supervision provides dense, local geometric cues that are more robust than direct 3D supervision and capture perceptually critical fine-scale features such as wrinkles and subtle curvature variations. In contrast, landmark supervision enforces semantic alignment by anchoring key structural regions---eyes, mouth, and facial contour---thus stabilizing global shape and promoting coherent edge flow during optimization.
For normal prediction, we adopt the base model of Garcia et al.~\cite{garcia2025fine} and fine-tune it with an additional 100K synthetic samples to enhance local fidelity and normal consistency. For landmark detection, we employ an RTMPose-style network~\cite{jiang2023rtmpose} comprising a CSPNeXt backbone~\cite{lyu2022rtmdet} and SimCC head~\cite{li2022simcc}, trained on 100K synthetic renderings and 50K FFHQ~\cite{karras2019style} images. Training configurations and additional implementation details are provided in our supplementary material.

\noindent \textbf{Normal Consistency.} 
To ensure that the reconstructed mesh preserves high-frequency geometric features that are critical to identity, we introduce an $\ell_1$ normal consistency loss:
$
\mathcal{L}_{\text{n}} =  \|\,\mathcal{N}(I) - \hat{\mathcal{N}}(V)\,\|_{1},
$
where $\mathcal{N}(I)$ denotes the per-pixel surface normals predicted from the input image, and $\hat{\mathcal{N}}(V)$ represents the rendered surface normals computed from the current 3D mesh with vertices $V$.
Supervising reconstruction with surface normals provides a rich geometric signal beyond classic point-wise correspondence, enabling the model to better capture identity-specific local curvature and fine-scale shape variations. This normal alignment encourages high-fidelity reproduction of individual facial structures, which is essential for faithful identity modeling.

\begin{figure}
    \centering
    \includegraphics[width=0.8\linewidth]{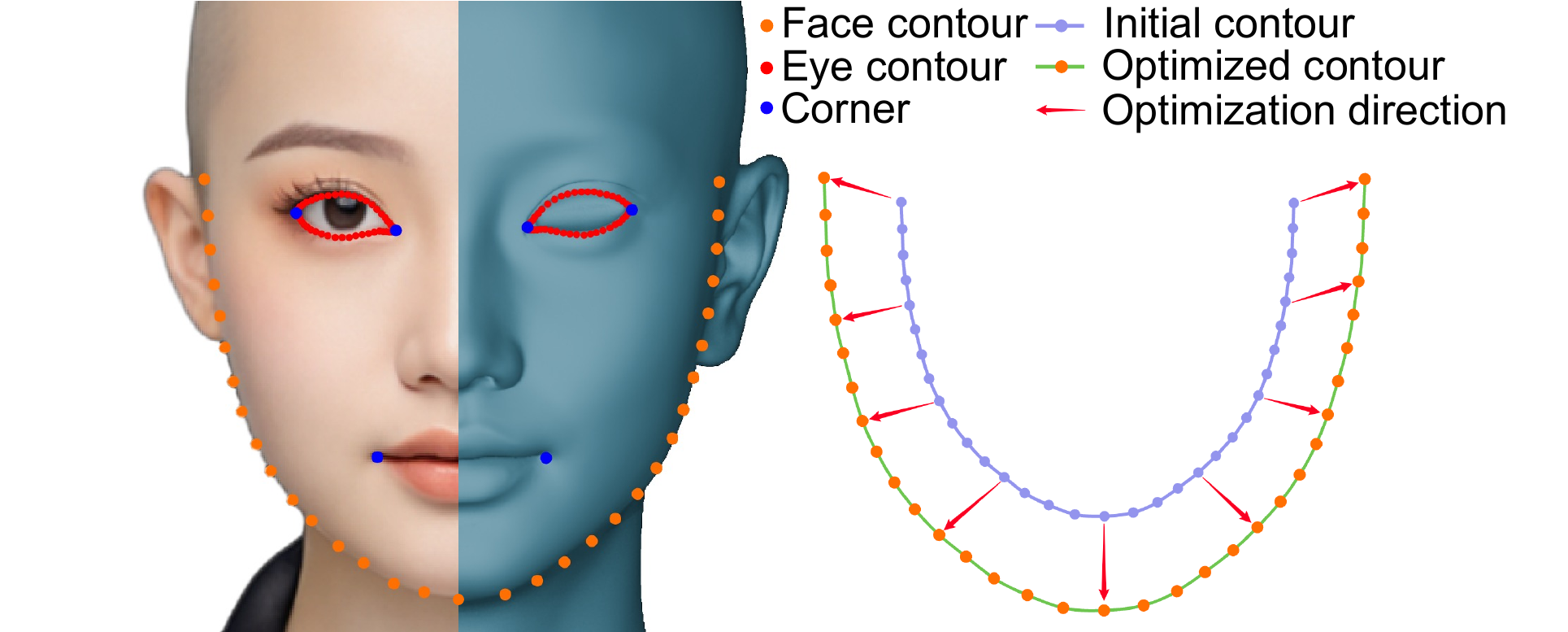}
    \vspace{-10pt}
    \caption{\textbf{Definition of our landmarks.} Besides classic corner landmarks, we introduce sequential point trajectories to better guide the reconstruction of identity-relevant features, such as eye shape and overall face contour.}
    \label{fig:landmark}
\end{figure}

\noindent \textbf{Landmark Alignment.}
As shown in~\figref{fig:landmark}, our landmark set includes 
(1) corner points of the eyes and mouth and 
(2) sequential landmarks along the eye and facial contours. 
The corner points provide sparse yet reliable anchors that constrain the global scale and orientation of the reconstruction, 
and their alignment is measured by a Mean Squared Error (MSE) loss. 
The contour landmarks offer denser supervision for capturing local geometric details and boundary consistency. 
Because the number of detected landmarks and projected mesh vertices along the contours may differ, 
we measure their correspondence using a one-sided Chamfer distance.

The overall landmark alignment loss is defined as:
\begin{equation}
\mathcal{L}_{\text{land}} =
\frac{1}{N_c} \sum_{i=1}^{N_c} 
\left\| \hat{\mathbf{u}}_i - \mathbf{u}_i \right\|_2^2
+ 
\frac{1}{|\hat{U}|} \sum_{\hat{\mathbf{u}} \in \hat{U}} 
\min_{\mathbf{u} \in U} 
\left\| \hat{\mathbf{u}} - \mathbf{u} \right\|_2^2,
\label{eq:land}
\end{equation}
where $N_c$ is the number of corner points, 
$\hat{\mathbf{u}}_i$ and $\mathbf{u}_i$ are the projected and detected 2D coordinates of the $i$-th corner, 
and $\hat{U}$ and $U$ denote the sets of projected and detected contour landmarks.

\subsubsection{Topological Consistency Enforcement}\label{subsec:stiffness}
Normal and landmark losses preserve global identity but often produce locally inconsistent mesh structures, which are unsuitable for industry-grade head meshes. To ensure geometric plausibility during deformation, we introduce two complementary topological consistency constraints:
$
\mathcal{L}_{\text{con}} = \lambda_{\text{GCC}}\mathcal{L}_{\text{GCC}} + \lambda_{\text{conf}}\mathcal{L}_{\text{conf}},
$
where $\mathcal{L}_{\text{GCC}}$ enforces curvature consistency and $\mathcal{L}_{\text{conf}}$ encourages local angle preservation. Together, these terms maintain intrinsic surface properties and suppress non-physical distortions during optimization.

\noindent \textbf{Edge-to-Vertex Gaussian Curvature Constraint (EV-GCC).} 
Gaussian curvature constraints are commonly used in mesh deformation~\cite{meyer2003discrete,tajdari2022feature} to preserve intrinsic geometry. However, vertex-based curvature computed via angle deficits suffers from irregular neighborhoods and poor parallelization, while edge-based dihedral-angle formulations are difficult to integrate with vertex-level supervision in differentiable frameworks.
To reconcile these trade-offs, we introduce an edge-to-vertex curvature formulation that redistributes edge-level dihedral angles to adjacent vertices using normalized weights. This enables vertex-level curvature supervision while preserving the locality and efficiency of edge-based processing. For each edge $e_{ij}$, the dihedral angle is computed as
$\theta_{ij}^e = \arccos(\mathbf{n}_1 \cdot \mathbf{n}_2),$

where $\mathbf{n}_1$ and $\mathbf{n}_2$ denote the normals of the adjacent faces sharing $e_{ij}$. The curvature at vertex $i$ is accumulated as
\begin{equation}
c_i = \sum_{j=1}^{n_{c_i}} \omega_{i \rightarrow ij}\,\theta_{ij}^e, 
\quad 
\omega_{i \rightarrow ij} = 
\frac{I_{i \rightarrow ij}}{\sum_k I_{i \rightarrow ik}},
\end{equation}
where $I_{i \rightarrow ij}$ represents the importance weight of edge $e_{ij}$. The curvature consistency loss is defined as
$\mathcal{L}_{\text{GCC}} = \sum_{i=1}^{N_v} \left| c_i^s - c_i^t \right|,$
with $c_i^s$ and $c_i^t$ denoting curvature values at vertex $i$ on the source and target meshes, respectively.

\noindent\textbf{Angle Preservation Constraint.}
We further enforce local conformality using an angle-preservation term inspired by mesh parameterization and UV unwrapping~\cite{gu2003global,yoshiyasu2014conformal,levy2023least}. This constraint penalizes deviations between corresponding internal triangle angles:
$
\mathcal{L}_{\text{conf}} = \sum_{i \in \Omega_\theta} \| \theta_i^s - \theta_i^t \|,
$
where $\theta_i^s$ and $\theta_i^t$ denote the internal angles at the $i$-th triangle corner in the source and target meshes. To prioritize regions undergoing large deformations, we rank the angular discrepancies and retain the top $\Omega_\theta$ entries for loss computation.

Together, EV-GCC and conformal constraints enforce topological coherence and geometric realism, significantly improving surface stability under large facial deformations.

\subsubsection{Fine Artifacts Suppression}\label{subsec:fine_artifacts}
While topological consistency constraints preserve intrinsic geometry and surface regularity, subtle artifacts can persist in visually sensitive regions (e.g., lips and eyes) due to high deformability. We therefore introduce fine-level regularization to enforce local orientation, contour smoothness, and optimization stability:
\begin{equation}
\mathcal{L}_{\text{reg}} = 
\lambda_{\text{flip}}\mathcal{L}_{\text{flip}} +
\lambda_{\text{curve}}\mathcal{L}_{\text{curve}} +
\lambda_{\text{disp}}\mathcal{L}_{\text{disp}},
\end{equation}
where $\mathcal{L}_{\text{flip}}$ prevents normal flipping, $\mathcal{L}_{\text{curve}}$ smooths eyeline contours, and $\mathcal{L}_{\text{disp}}$ stabilizes vertex displacements.

\noindent \textbf{Face-Flip Constraint.}
Flipping artifacts in highly deformable facial regions such as the lips and oral cavity can produce visually implausible shading and animation artifacts. 
To suppress such effects, we introduce a normal regularization that penalizes large deviations in face orientations between corresponding source and target meshes:
$
\mathcal{L}_{\text{flip}} = \sum_{i \in \Omega_{\vec{n}}} \left\lVert \vec{n}_i^s - \vec{n}_i^t \right\rVert,
$
where \( \vec{n}_i^s \) and \( \vec{n}_i^t \) denote the unit normals of the \(i\)-th face in the source and target meshes. 
Similar to $\mathcal{L}_{\text{conf}}$, the top $\Omega_{\vec{n}}$ faces with the largest normal deviations are selected to focus supervision on regions most prone to inversion.
This constraint preserves consistent surface orientation and suppresses normal flips that disrupt local topology.

\noindent \textbf{Eyeline Curve Flow Constraint.}
The eyes are among the most perceptually critical facial regions, yet their fine geometric structures often lead to jagged contours during deformation. 
We introduce a weak regularizer that promotes smooth directional flow along key curves (e.g., the upper eyelid crease and outer eye contour) while remaining tolerant to subject-specific variations. 
As an example, for the upper eye outline defined by an ordered vertex sequence $\{v_1, \dots, v_n\}$, we compute local direction vectors:
\begin{equation}
\vec{d}_i = \tfrac{1}{2} \left( 
\frac{\vec{v}_{i+1} - \vec{v}_i}{\|\vec{v}_{i+1} - \vec{v}_i\|} + 
\frac{\vec{v}_{i+2} - \vec{v}_{i+1}}{\|\vec{v}_{i+2} - \vec{v}_{i+1}\|}
\right),
\label{eq:flow_dir}
\end{equation}
for $i = 1, \dots, n-2$. 
The turning smoothness feature is then defined as the normalized angle change between consecutive directions:
\begin{equation}
f_i = \frac{1}{\|\vec{v}_{i+1} - \vec{v}_i\|} \,
\arccos \!\left( 
\frac{\vec{d}_i \cdot \vec{d}_{i+1}}{\|\vec{d}_i\| \, \|\vec{d}_{i+1}\|}
\right).
\label{eq:flow_feature}
\end{equation}
We measure the $\ell_1$ discrepancy between the flow features of the source and target curves:
$
\mathcal{L}_{\text{curve}} = \sum_{i=1}^{n-3} \left| f_i^s - f_i^t \right|.
\label{eq:flow_loss}
$
This term enforces smooth curvature transitions along delicate contours, suppressing unnatural jaggedness around the eyes.

\noindent \textbf{Vertex Displacement Regularization.} 
We further impose a regularization on vertex displacements $\mathcal{L}_{\text{disp}} = \sum_{i=1}^{N_v} \|\Delta \mathbf{v}_i\|$ to stabilize the deformation optimization process. This term mitigates cumulative geometric drift by encouraging smooth shape evolution, resulting in more stable convergence and fewer local artifacts.

\subsection{Texture Extraction} \label{subsec:texture}

Following geometry reconstruction, we generate appearance maps via a UV-space texture pipeline as illustrated in~\figref{fig:texture}. To bridge the resolution gap between the input image and the target UV map, we first upsample the input using a Real-ESRGAN-based super-resolution module \cite{wang2021realesrgan}. Subsequently, we employ an intrinsic decomposition model \cite{liang2025diffusion} to extract the albedo, effectively decoupling shading effects while preserving fine-grained details. Meanwhile, a heuristic normal map is generated in the image space. Both the albedo and the normal map are projected directly onto the UV space. Unlike per-vertex color baking, this rasterization strategy preserves high-frequency details. Finally, to address missing regions caused by self-occlusion, we utilize the Dual-Mask Poisson Blending method to inpaint the texture, ensuring gradient consistency and color coherence in the final UV map. We highlight two components that deviate from conventional pipelines:

\noindent\textbf{Normal Computation.}
Neural network–predicted normals often suffer from scale ambiguity and overly smoothed high-frequency details. To address this, we adopt a gradient-based strategy~\cite{ratz_2017_normalmapgenerator}. Specifically, the Sobel operator is applied to both fine- and coarse-resolution facial images to compute horizontal and vertical gradients $(g_{i,j}^x, g_{i,j}^y)$. The tangent-space normal is defined as $\mathbf{n}_{i,j} \propto (g_{i,j}^x,\, g_{i,j}^y,\, 1)$. The two normal maps are then fused using a soft-light operation, which preserves high-frequency geometric cues while maintaining smooth global structure, resulting in stable and detailed surface normals.

\noindent\textbf{Dual-Mask Poisson Blending.}
Since single-view texture predictions are inherently incomplete due to occlusions, we complete them by blending with template textures $\widetilde{\mathcal{A}}$. To avoid artifacts and tone shifts that arise from direct Poisson blending, we employ a dual-mask blending scheme. A soft parsing mask guides smooth transitions in interior regions, while a hard boundary mask preserves precise seam constraints. We derive the hard mask directly from a facial parsing map~\cite{face-parsing} and generate the soft mask by applying a $3\times3$ erosion kernel. This dual-mask approach produces seamless transitions while maintaining consistent global appearance throughout the UV map.

\section{Experiments}
\label{sec:experiments}

\subsection{Experimental settings}

\textbf{Settings.}
In the main quantitative experiments, we use a rigged template following the MetaHuman mesh topology~\cite{EpicGames2021Metahuman} for illustration. The template comprises \textit{24{,}049} vertices, \textit{808} joints, and \textit{147} rig controllers, supporting high-fidelity geometric deformation. We implement our method in PyTorch~\cite{paszke2019pytorch} and use the differentiable rasterizer from PyTorch3D~\cite{ravi2020pytorch3d}. Optimization hyperparameters and learning-rate schedules are provided in the supplementary material. All experiments are performed on a workstation with an NVIDIA Tesla V100 (32\,GB) GPU; CPU-only runs use an Intel Xeon Platinum 8163 processor. On this hardware, our full pipeline takes about 4.5 minutes per input image, including 2.5 minutes for geometry reconstruction and 2 minutes for texture extraction. Our reconstruction assumes a near-neutral, unoccluded face image. For samples with large pose or significant occlusion, we first apply a fine-tuned FLUX.1 Kontext~\cite{labs2025flux1kontextflowmatching} preprocessing model to frontalize the face and remove major occluders, as shown in the inset of~\figref{fig:teaser}. This preprocessing is used only to satisfy the input assumption; when it is needed, we apply the same preprocessed input to all compared methods for fairness. Additional preprocessing details are provided in the supplementary material.

\noindent\textbf{Baselines.} We compare our method against widely used single-image face reconstruction methods and classical mesh registration approaches. For reconstruction, we include FaceScape~\cite{yang2020facescape}, DECA~\cite{zielonka2022towards}, MICA~\cite{zielonka2022towards}, and Pixel3DMM~\cite{giebenhain2025pixel3dmm}, all of which rely on low-dimensional parametric spaces (e.g., FaceScape or FLAME~\cite{li2017learning} shape bases) and output relatively low-resolution meshes. For registration baselines, we evaluate NR-ICP~\cite{amberg2007optimal} and ACAP~\cite{yoshiyasu2014conformal} under the same template topology, and include FLAME as a representative parametric registration method.

\noindent \textbf{Datasets.} Our experiments are evaluated on two datasets. To ensure access to accurate ground truth, we construct a synthetic dataset by combining 66 publicly available head models~\cite{EpicGames2021Metahuman} with 40 artist-created high-quality assets. In addition, we collect 158 in-the-wild images from public web sources, academic datasets~\cite{karras2019style, liu2015faceattributes}, and image generation tools, covering a wide variety of conditions and styles. These evaluation images span diverse human subjects across skin tones, as well as non-photorealistic domains such as anime characters. This data composition supports broad qualitative and quantitative evaluation of our method.

\noindent \textbf{Evaluation Metrics.} 
We evaluate our method using standard geometric metrics, including Root Mean Square Error (RMSE) and Normal Consistency (NC), as well as ArcFace similarity, which measures identity preservation--an aspect that is often more important than pure geometric accuracy for face reconstruction, since faithful identity retention is essential for downstream binding and animation workflows. For a fair comparison of identity preservation, we render all reconstructed heads using PyTorch3D under the same front-view camera, material, and lighting configuration, and compute the ArcFace~\cite{deng2019arcface} cosine similarity between the rendered images and the input image. To ensure fair geometric comparison across all methods, we uniformly resample all meshes to a fixed number of points and evaluate errors within a masked facial region. Further implementation details for the metric computation are provided in the supplementary material.

\subsection{Results}

\begin{table}[t]
  \centering
  \small
  \setlength{\tabcolsep}{3pt}
  \setlength{\aboverulesep}{0pt}
  \setlength{\belowrulesep}{0pt}
  \begin{tabular*}{\linewidth}{@{\extracolsep{\fill}}l|cccc|c@{}}
    \toprule
    \multirow{2}{*}[-0.7ex]{Method}
      & \multicolumn{4}{c|}{Synthetic}
      & \multicolumn{1}{c}{In-the-wild} \\
    \cmidrule(lr){2-5} \cmidrule(lr){6-6}
      & RMSE$\,\downarrow$ & NC$\,\uparrow$ & ArcFace$\,\uparrow$ & ArcFace$^*\,\uparrow$ & ArcFace$\,\uparrow$ \\
    \midrule
    FaceScape & 0.0318 & 0.9609 & 0.1765 & 0.3173 & 0.1341 \\
    DECA & 0.0402 & 0.9706 & 0.1834 & 0.4359 & 0.1014 \\
    MICA & 0.0474 & 0.9711 & 0.1621 & 0.3980 & 0.1213 \\
    Pixel3DMM & 0.0630 & 0.9704 & 0.1876 & 0.4332 & 0.1472 \\
    Ours & \textbf{0.0159} & \textbf{0.9798} & \textbf{0.3770} & \textbf{0.6322} & \textbf{0.4140} \\
    \bottomrule
  \end{tabular*}
  \caption{\textbf{Comparison of single-image head reconstruction methods 
  on \emph{Synthetic} and \emph{In-the-wild} datasets.} 
  $^*$ denotes the ArcFace score computed on ground-truth renderings under the same camera, lighting, and diffuse gray material.}
  \label{tab:recon_tab_corr}
\end{table}

\begin{table}[t]
  \centering
  \small
  \setlength{\tabcolsep}{3pt}
  \setlength{\aboverulesep}{0pt}
  \setlength{\belowrulesep}{0pt}
  \begin{tabular*}{\linewidth}{@{\extracolsep{\fill}}l|cccc@{}}
    \toprule
    Method & RMSE$\,\downarrow$ & NC$\,\uparrow$ & ArcFace$\,\uparrow$ & ArcFace$^*\,\uparrow$ \\
    \midrule
    FLAME & 0.0215 & 0.9755 & 0.2192 & 0.4915 \\
    NR-ICP & \textbf{0.0013} & 0.9944 & 0.4196 & 0.8196 \\
    ACAP & 0.0028 & 0.9910 & 0.3654 & 0.6793 \\
    Ours & 0.0019 & \textbf{0.9975} & \textbf{0.4397} & \textbf{0.9421} \\
    \bottomrule
  \end{tabular*}
  \caption{\textbf{Comparison with 3D face registration methods.}
  $^*$ denotes the ArcFace score computed on ground-truth renderings under the same camera, lighting, and diffuse gray material.}
  \label{tab:reg_tab_corr}
\end{table}

\begin{figure*}[!t]
\centering

\setlength{\abovecaptionskip}{2pt}
\setlength{\belowcaptionskip}{2pt}

\begin{minipage}{\textwidth}
  \centering
  \includegraphics[width=\linewidth]{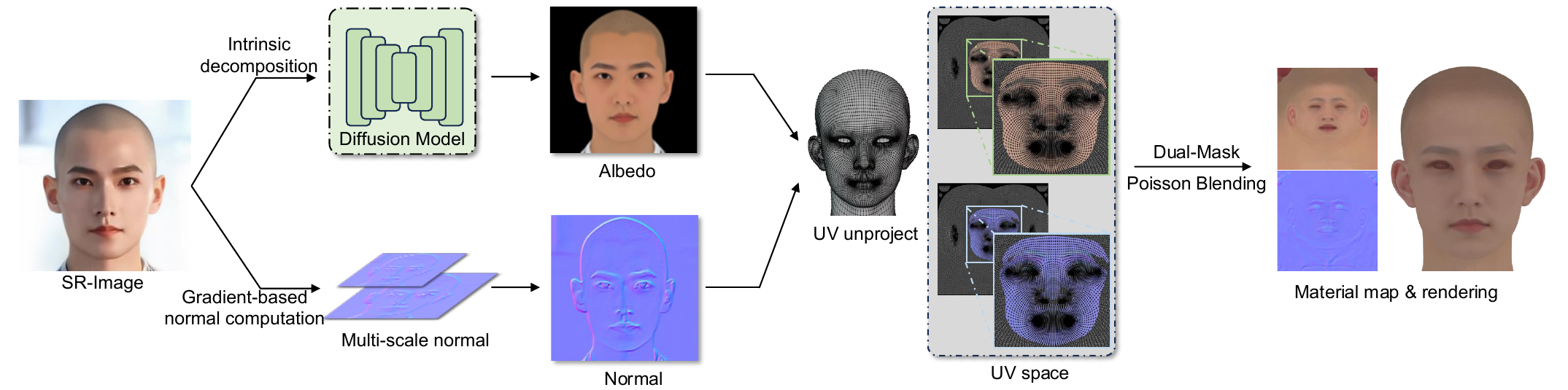}
  \vspace{-12pt}
  \caption{\textbf{Pipeline of texture extraction.}
    The input image is first super-resolved and decomposed to obtain albedo, while a heuristic normal map is estimated in image space. Both albedo and normal maps are then projected onto the UV map via rasterization, preserving high-frequency details. Finally, missing regions due to self-occlusion are completed using Dual-Mask Poisson Blending to ensure color and gradient consistency.
  }
  \label{fig:texture}
\end{minipage}

\vspace{4pt}

\begin{minipage}{\textwidth}
  \centering
    \begin{minipage}{\linewidth}
        \centering
        \small
      \begin{minipage}{0.08\linewidth}
        \centering
         Input
      \end{minipage}
      \begin{minipage}{0.08\linewidth}
        \centering
         Ours
      \end{minipage}
      \begin{minipage}{0.08\linewidth}
        \centering
         Pixel3DMM
      \end{minipage}
      \begin{minipage}{0.08\linewidth}
        \centering
         MICA
      \end{minipage}
      \begin{minipage}{0.08\linewidth}
        \centering
         DECA
      \end{minipage}
      \begin{minipage}{0.08\linewidth}
        \centering
         FaceScape
      \end{minipage}
      \begin{minipage}{0.08\linewidth}
        \centering
         Input
      \end{minipage}
      \begin{minipage}{0.08\linewidth}
        \centering
         Ours
      \end{minipage}
      \begin{minipage}{0.08\linewidth}
        \centering
         Pixel3DMM
      \end{minipage}
      \begin{minipage}{0.08\linewidth}
        \centering
         MICA
      \end{minipage}
      \begin{minipage}{0.08\linewidth}
        \centering
         DECA
      \end{minipage}
      \begin{minipage}{0.08\linewidth}
        \centering
         FaceScape
      \end{minipage}
    \end{minipage}
  \begin{minipage}{0.5\linewidth}
    \includegraphics[width=\linewidth]{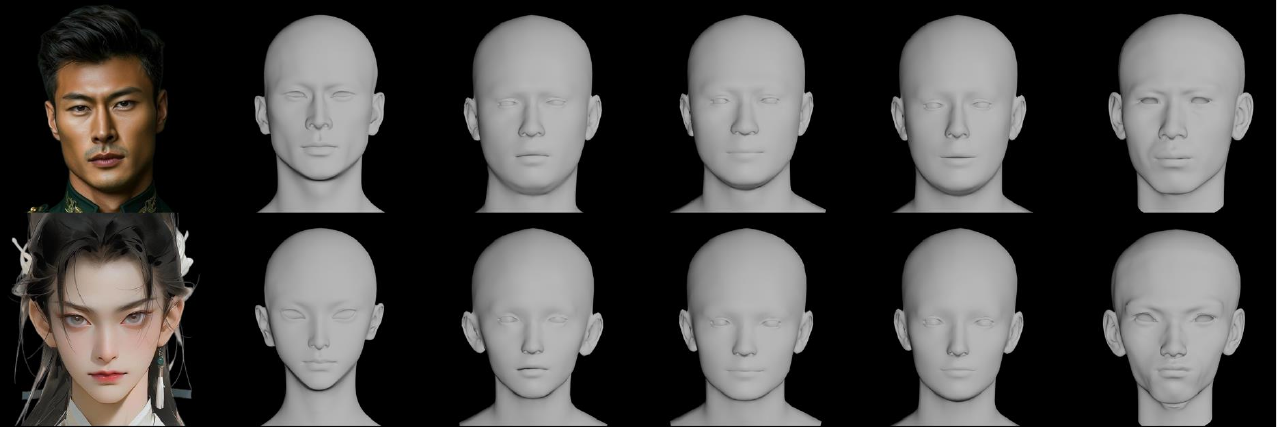}
  \end{minipage}%
  \begin{minipage}{0.5\linewidth}
    \includegraphics[width=\linewidth]{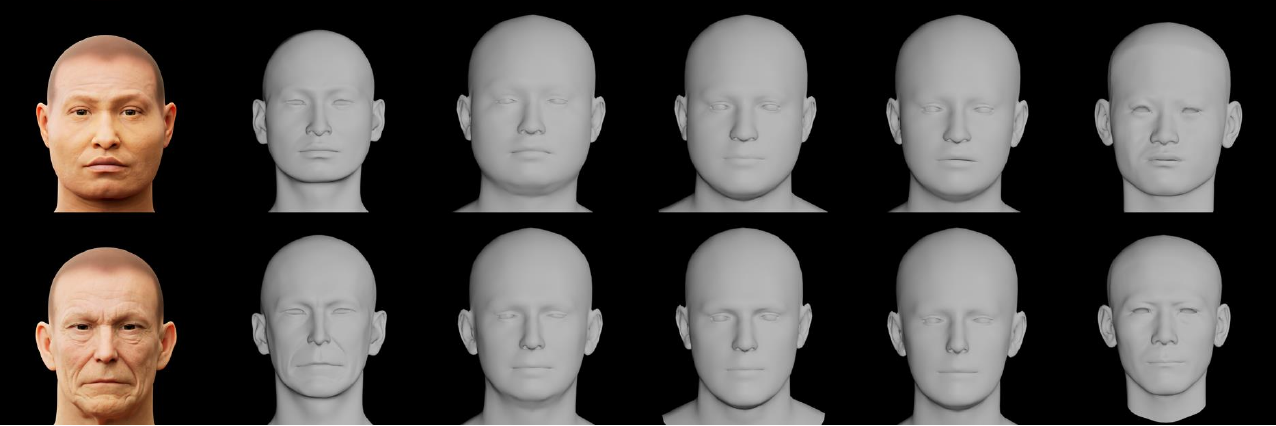}
  \end{minipage}

  \caption{\textbf{Comparison of single-image head reconstruction methods on \emph{in-the-wild} and \emph{synthetic} datasets.}
    The two rows on the left show results on in-the-wild images, while the two rows on the right present reconstructions on synthetic data. Across both settings, our method produces noticeably more realistic and identity-faithful heads than existing methods.
  }
  \label{fig:face_recon}
\end{minipage}

\vspace{6pt}

\begin{minipage}{\textwidth}
\centering

\begin{minipage}[t]{0.49\textwidth}
  \centering

  \begin{minipage}{0.196\linewidth}
    \centering
     Input
  \end{minipage}%
  \begin{minipage}{0.2\linewidth}
    \centering
     Ours
  \end{minipage}%
  \begin{minipage}{0.22\linewidth}
    \centering
     ACAP
  \end{minipage}%
  \begin{minipage}{0.22\linewidth}
    \centering
     NR-ICP
  \end{minipage}%
  \begin{minipage}{0.22\linewidth}
    \centering
     FLAME
  \end{minipage}

  \includegraphics[width=\linewidth]{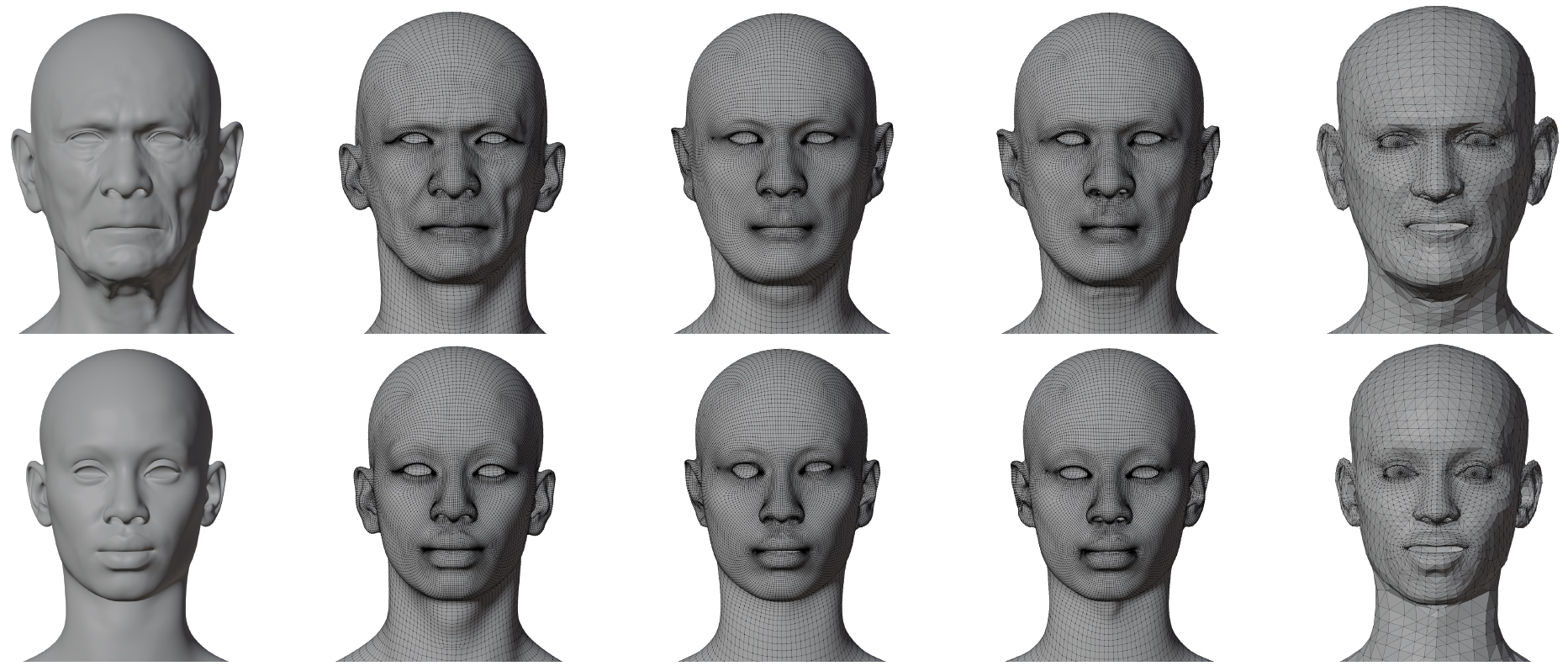}
  \vspace{-10pt}
  \caption{\textbf{Comparison with registration-based methods.}
    From left to right in each row: ground-truth rendering, our method, ACAP~\cite{yoshiyasu2014conformal}, NR-ICP~\cite{amberg2007optimal}, and FLAME~\cite{li2017learning}.
    Benefiting from our coarse-to-fine framework and geometry-aware regularization, our method reconstructs head geometry with topology regularity comparable to registration-based approaches.
  }
  \label{fig:3d_cmp_new}

  \vspace{6pt}

  \includegraphics[width=\linewidth]{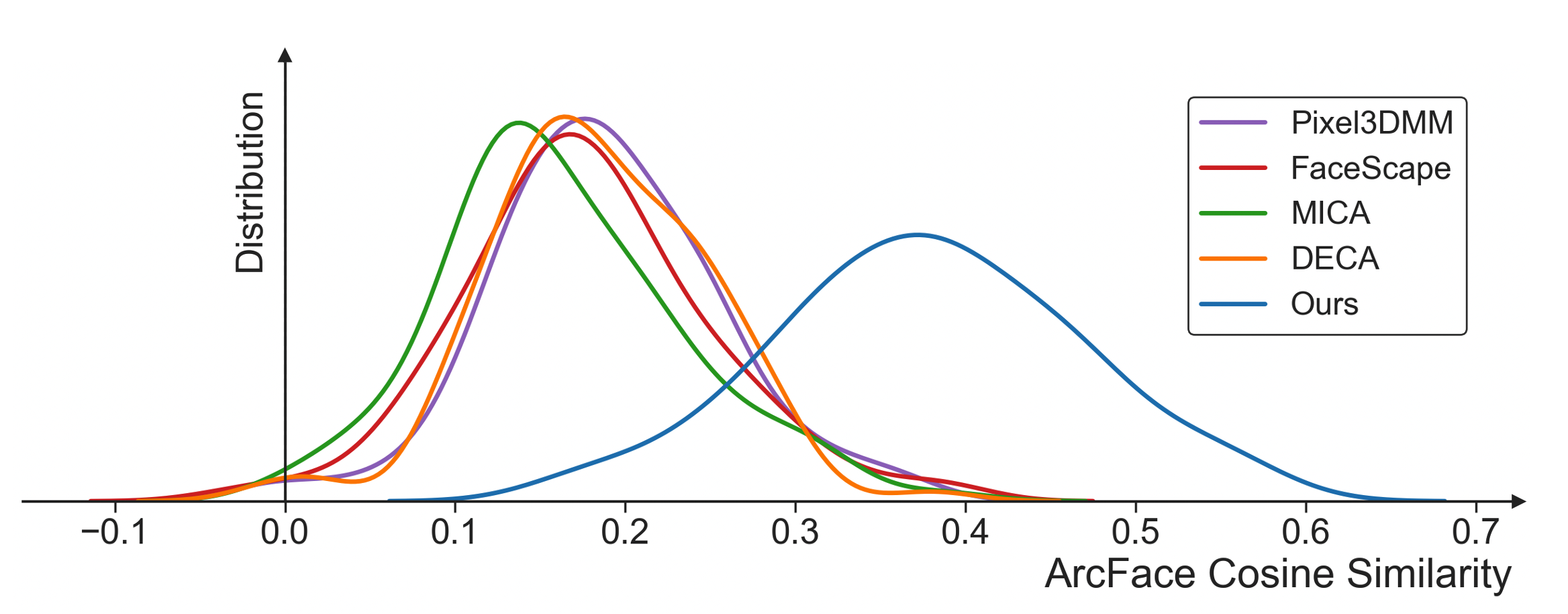}
  \vspace{-12pt}
  \caption{
    \textbf{ArcFace score distribution on the synthetic dataset.}
    Our method ({{\color{blue}blue}}) yields consistently higher ArcFace scores than other single-image approaches on this evaluation set.
  }
  \label{fig:cum_sim}

\end{minipage}
\hfill
\begin{minipage}[t]{0.49\textwidth}
  \centering

  \begin{minipage}{0.24\linewidth}
    \centering
     Ours
  \end{minipage}%
  \begin{minipage}{0.26\linewidth}
    \centering
     w/o $\mathcal{L}_{\text{GCC}}$
  \end{minipage}%
  \begin{minipage}{0.26\linewidth}
    \centering
     w/o $\mathcal{L}_{\text{conf}}$
  \end{minipage}%
  \begin{minipage}{0.24\linewidth}
    \centering
     w/o coarse stg.
  \end{minipage}

  \includegraphics[width=\linewidth]{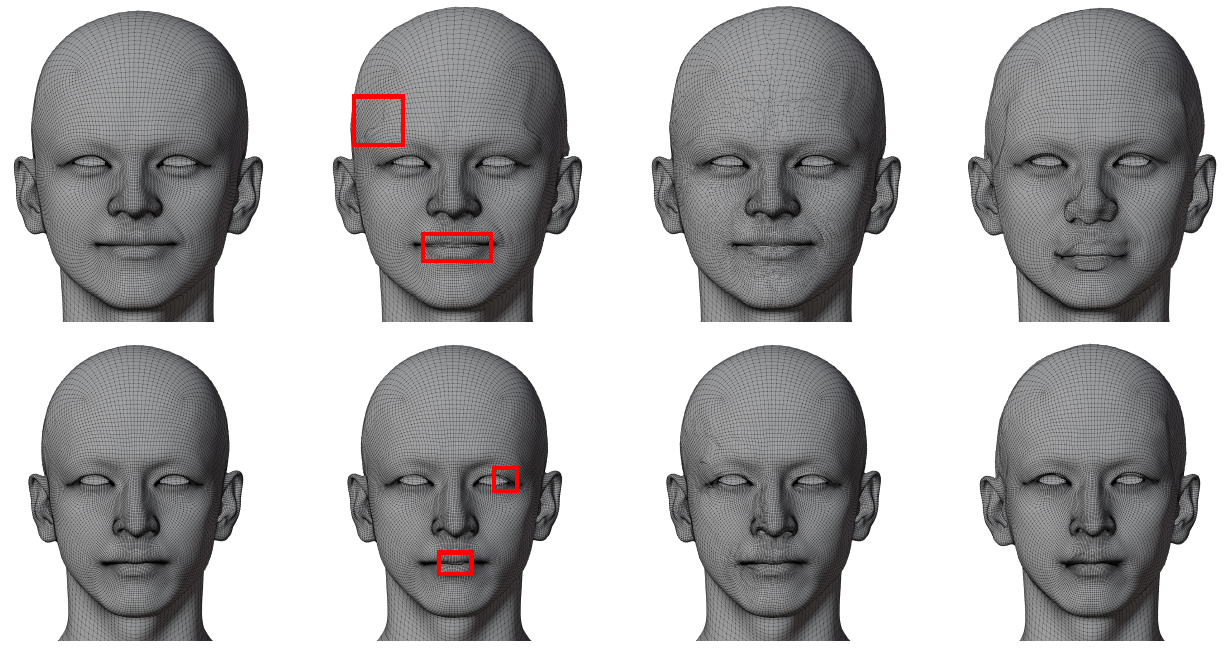}
  \vspace{-12pt}
  \caption{
    \textbf{Ablation study of our regularizations and coarse-to-fine framework.}
    From left to right: our method, results without Edge-to-Vertex Gaussian Curvature Constraint, results without angle preservation constraints, and direct vertex-level optimization. The local artifacts caused by removing EV-GCC are highlighted with red boxes for clarity, making it easier to identify the artifacts introduced by this modification.
  }
  \label{fig:ablation_new}

  \vspace{6pt}

  \begin{minipage}{0.19\linewidth}
    \centering
     Input
  \end{minipage}%
  \begin{minipage}{0.2\linewidth}
    \centering
     Recon.
  \end{minipage}%
  \begin{minipage}{0.21\linewidth}
    \centering
     Layout
  \end{minipage}%
  \begin{minipage}{0.2\linewidth}
    \centering
     Recon.
  \end{minipage}%
  \begin{minipage}{0.19\linewidth}
    \centering
     Layout
  \end{minipage}

  \includegraphics[width=\linewidth]{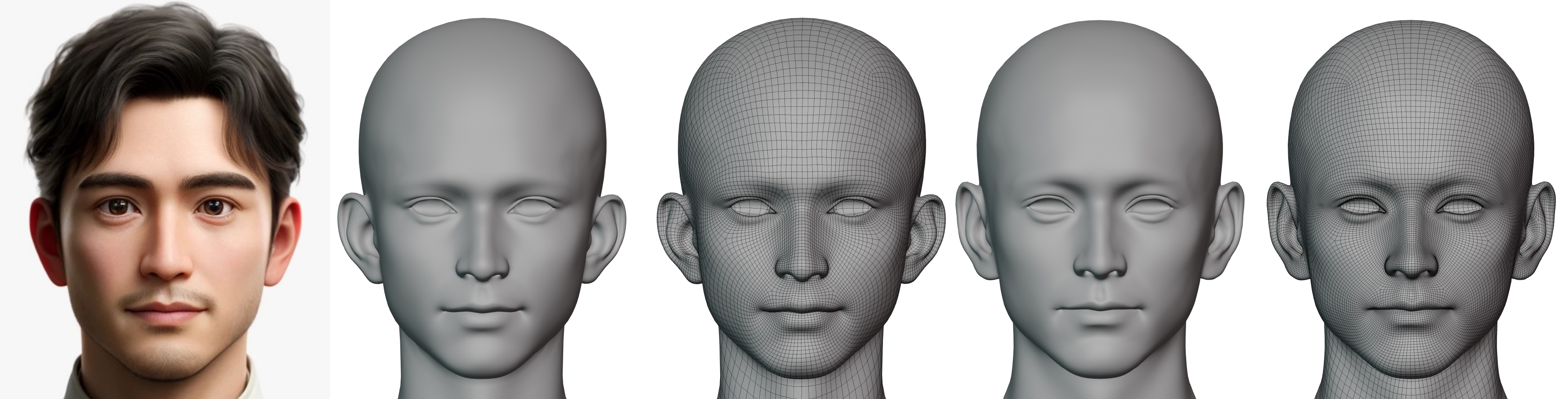}
  \vspace{-10pt}
  \caption{
    \textbf{Qualitative evaluation on different head-mesh topologies.}
    From left to right: input image, reconstruction with a simplified topology, its mesh layout, reconstruction with MetaHuman topology, and its mesh layout. Our method preserves identity and produces clean, semantically consistent topology in both settings.
  }
  \label{fig:different_topo}

\end{minipage}

\end{minipage}

\end{figure*}

\begin{figure*}
  \centering
  \vspace{-5pt}
  \includegraphics[width=\textwidth]{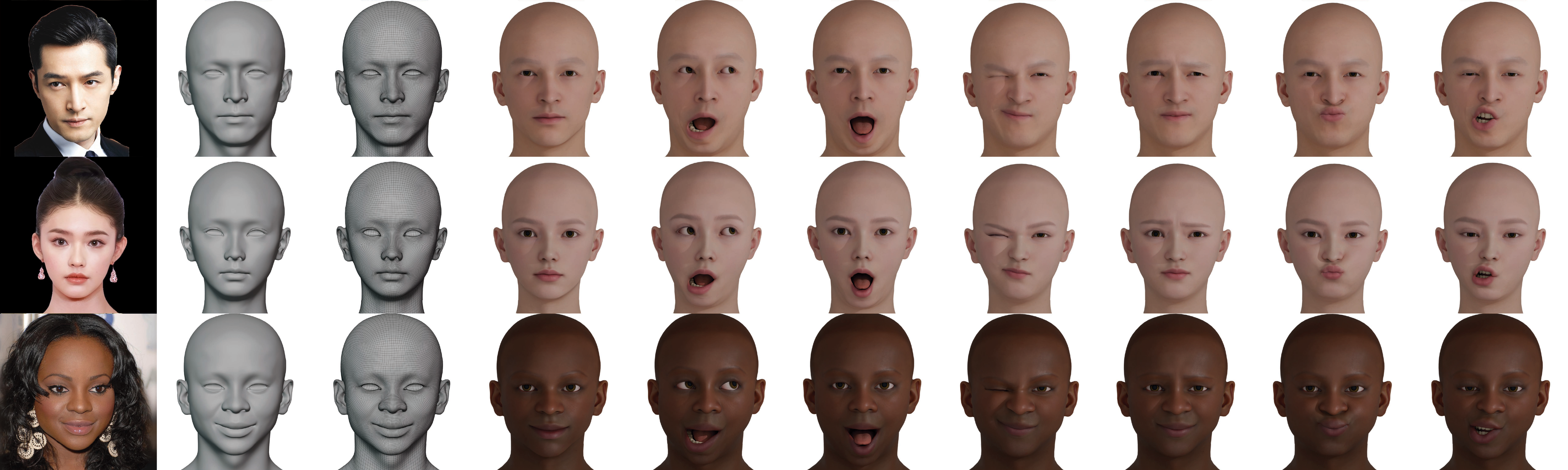}
  \vspace{-20pt}
    \caption{
    \textbf{More reconstruction and downstream deformation results produced by our method.} From left to right: the input image, reconstructed geometry, mesh topology, textured reconstruction, and a sequence of extreme deformation tests. These examples show that our reconstructed meshes remain stable in challenging downstream deformation tests.
}
\vspace{-15pt}
    \label{fig:animation_full}
\end{figure*}

As shown in~\figref{fig:teaser} and \figref{fig:animation_full}, our method produces high-fidelity 3D reconstructions with strong identity preservation, faithfully recovering person-specific geometric structure. The reconstructed meshes also exhibit stable behavior after downstream binding in animation tests while maintaining identity consistency.
We further validate our framework on two different head-mesh topologies in~\figref{fig:different_topo} and obtain industry-grade reconstructions with strong identity preservation in both settings. The consistent behavior across these topology configurations provides empirical support that our method is not tied to a specific template topology.
Beyond these qualitative observations, we evaluate our approach through quantitative and qualitative comparisons against recent state-of-the-art methods.

\noindent \textbf{Single-image Face Reconstruction.}
We assess our method on both synthetic renderings and in-the-wild images. On synthetic data, we compare against representative 3DMM-based models, including FaceScape, DECA, MICA, and Pixel3DMM. As shown in~\figref{fig:face_recon}, among existing parametric models only FaceScape produces roughly comparable facial proportions, but its limited shape space often introduces noticeable artifacts, such as overly pronounced brow ridges and jawlines. In contrast, our method achieves accurate, stable reconstruction with substantially better identity retention, reflected by significantly higher ArcFace cosine similarity scores in~\tabref{tab:recon_tab_corr}.
To evaluate generalization, we conduct experiments on in-the-wild images. Our method consistently outperforms all baselines across diverse lighting, pose, and appearance conditions. The ArcFace similarity distributions in~\figref{fig:cum_sim} further show that our reconstructed identities remain more faithful, even under uncontrolled imaging conditions.

\begin{figure*}
  \centering
        \begin{minipage}{0.1\linewidth}
        \centering \small Input
      \end{minipage}%
      \begin{minipage}{0.2\linewidth}
        \centering \small Ours
      \end{minipage}%
      \begin{minipage}{0.2\linewidth}
        \centering \small DreamFace
      \end{minipage}%
        \begin{minipage}{0.1\linewidth}
        \centering \small Input
      \end{minipage}%
      \begin{minipage}{0.2\linewidth}
        \centering \small Ours
      \end{minipage}%
      \begin{minipage}{0.2\linewidth}
        \centering \small DreamFace
      \end{minipage}
    {\includegraphics[width=\linewidth]{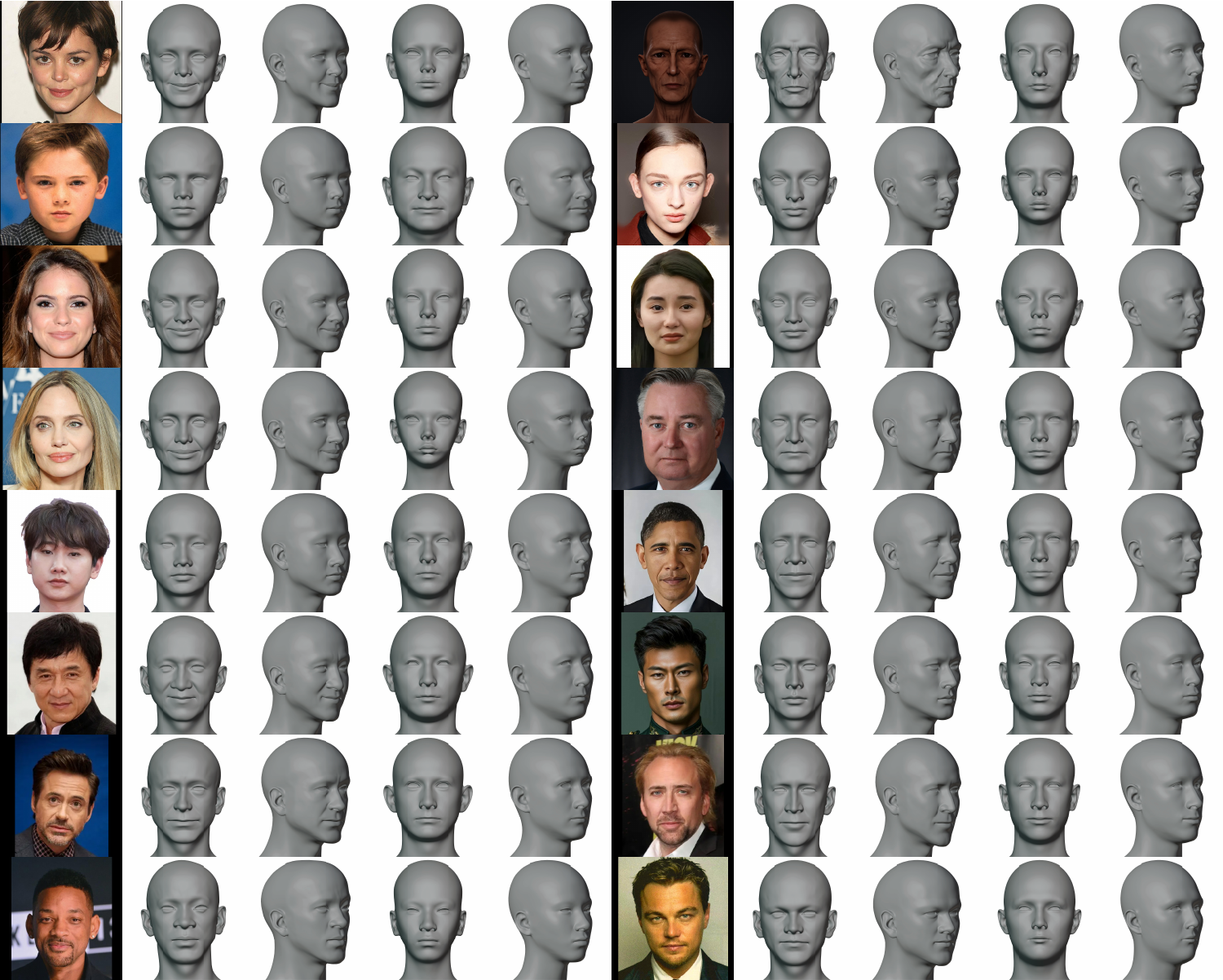}}
    \vspace{-20pt}
    \caption{
\textbf{Comparison with DreamFace.}
Each row from left to right: input image, our reconstruction, and DreamFace reconstruction, repeated for two examples. Our method better preserves global facial structure and fine-grained details in these examples.
}
    \label{fig:cmp_dreanface}
\end{figure*}

\noindent\textbf{Comparison with DreamFace~\cite{zhang2023dreamfaceprogressivegenerationanimatable}.}
Since DreamFace is closed-source and does not provide a batch API, we report ArcFace measurements only on our in-the-wild dataset. Using the official web interface provided by \cite{Hyper3DChatAvatar}, DreamFace achieves an ArcFace score of \textit{0.1821}, whereas our method reaches \textit{0.4140}. As shown in~\figref{fig:cmp_dreanface}, DreamFace preserves identity less faithfully, often failing to recover fine-grained facial details and, in some cases, even struggling to reproduce the coarse facial structure (e.g., rows 3 and 4). In contrast, our method reliably reconstructs both the global facial geometry and subtle high-frequency cues, including fine surface details such as wrinkles.

\noindent\textbf{Scan-to-fitting Face Registration.}
We also compare our fitting accuracy with classical optimization-based registration methods (NR-ICP and ACAP) and the parametric FLAME model. For this controlled comparison, our method uses rendered normals as input, whereas the other methods directly utilize the ground-truth model. Unlike learned registration baselines that require scan datasets for model fitting, our optimization stage does not require subject-specific scan training data. As illustrated in~\figref{fig:3d_cmp_new}, our method surpasses FLAME in 3D fitting accuracy while maintaining semantic consistency and producing clean, regular topology. Quantitative results reported in~\tabref{tab:reg_tab_corr} show that our method reaches accuracy comparable to classical optimization-based techniques while operating from only a \emph{single} input image. It preserves individual identity more faithfully and produces coherent edge flows, highlighting both the effectiveness and the generality of our registration design.

\noindent \textbf{Downstream Application.}
We further evaluate the usability of the reconstructed heads in downstream facial animation workflows. For this test, each reconstructed mesh is bound to an independent facial rig that is not used during reconstruction. As shown in~\figref{fig:animation_full}, our meshes maintain consistent topology and semantically aligned edge flow, enabling integration with standard facial rigs and expression controllers after binding. When driven by skeleton-based animation, the meshes deform smoothly and stably across large expressions and extreme poses in our test sequences, while preserving local details such as wrinkles and skin folds. We observe no visible tearing or texture distortion in the shown motion sequences. These results suggest that our method produces industry-grade head meshes well suited for downstream binding and animation.

\subsection{Ablations}

We conduct ablation studies to analyze the contribution of each key component in our optimization framework. As shown in~\figref{fig:pipeline} and \figref{fig:ablation_new}, removing any of these terms leads to noticeable degradation in geometric fidelity, surface smoothness, or topological regularity.

\noindent \textbf{Edge-to-Vertex Gaussian Curvature Constraint} mitigates surface artifacts caused by local optimization minima by enforcing curvature consistency with the standardized mesh. Without EV-GCC, the reconstructed surfaces exhibit high-frequency noise and irregular shading, as shown in the second column, indicating reduced geometric smoothness and local instability.

\noindent \textbf{Conformal Constraint} encourages locally smooth and consistent edge flows across the surface. When this term is removed, the edge orientations become irregular and lose global coherence, resulting in disrupted edge flow patterns and degraded surface regularity, as illustrated in the third column.

\noindent \textbf{Coarse-to-Fine Optimization Framework} provides stable initialization and progressively refines details, preventing the optimization from being trapped in suboptimal minima. Directly optimizing vertex positions in a single stage leads to unstable convergence, degraded geometry, and irregular topology that is unsuitable for downstream applications, as demonstrated in the last column.

\subsection{User Study}
We conducted a user study to assess perceptual quality in terms of global geometric fidelity, mesh layout usability, and overall preference. 22 professional technical artists evaluated 30 reconstructed faces produced by MICA, Pixel3DMM, and our method. On a five-point scale, our method achieved an average score of 3.6, substantially higher than Pixel3DMM (1.6) and MICA (1.42).
On the three-point topology-usability scale, our method obtained a score of 1.79, approaching industry-grade usability, whereas Pixel3DMM and MICA scored only 0.31 and 0.34. Overall, 95\% of participants ranked our method as the top-performing approach. These results indicate that our meshes achieve both stronger perceptual alignment and more regular edge flow, making them well-suited for downstream binding and animation workflows.
Additional details are provided in the supplementary material.

\section{Conclusion and Future Work}
We present a framework for automatic head mesh reconstruction from a single image. Our method produces highly identity-preserving head geometry while maintaining semantically meaningful, industry-grade edge flow and topology, making the reconstructed meshes suitable for downstream binding workflows. Despite these advantages, our current focus is primarily on geometric reconstruction. Material modeling and appearance fidelity remain important directions for production-quality digital humans.

\begin{acks}
We thank Kun Yang and all professional artists for their valuable contributions to this project.
\end{acks}

\bibliographystyle{ACM-Reference-Format}
\bibliography{main}
\clearpage

\end{document}